\documentclass[runningheads]{llncs}

\usepackage{eccv}

\usepackage{eccvabbrv}

\usepackage{graphicx}
\usepackage{booktabs}

\usepackage[accsupp]{axessibility}  % Improves PDF readability for those with disabilities.

\usepackage{hyperref}

\usepackage{relsize} %used to change the size of text
\usepackage{bbding} %for extra symbols \checkmark
\usepackage{pifont} %xmark
\newcommand{\cmark}{\ding{51}}%
\newcommand{\xmark}{\ding{55}}%
\usepackage[dvipsnames]{xcolor}
\usepackage{multirow}

\begin{document}

% ---------------------------------------------------------------
% TODO REVIEW: Replace with your title
\title{ABC Easy as 123: A Blind Counter \\ for Exemplar-Free Multi-Class  \\ Class-agnostic Counting} 

% TODO REVIEW: If the paper title is too long for the running head, you can set
% an abbreviated paper title here. If not, comment out.
\titlerunning{ABC Easy as 123}

% TODO FINAL: Replace with your author list. 
% Include the authors' OCRID for the camera-ready version, if at all possible.
\author{Michael Hobley \and
Victor Prisacariu}

% TODO FINAL: Replace with an abbreviated list of authors.
\authorrunning{Hobley and Prisacariu}
% First names are abbreviated in the running head.
% If there are more than two authors, 'et al.' is used.

% TODO FINAL: Replace with your institution list.
\institute{Active Vision Laboratory, University of Oxford \\
\email{[mahobley, victor]@robots.ox.ac.uk}
}
% % TODO FINAL: Replace with your institution list.
% \institute{Princeton University, Princeton NJ 08544, USA \and
% Springer Heidelberg, Tiergartenstr.~17, 69121 Heidelberg, Germany
% \email{lncs@springer.com}\\
% \url{http://www.springer.com/gp/computer-science/lncs} \and
% ABC Institute, Rupert-Karls-University Heidelberg, Heidelberg, Germany\\
% \email{\{abc,lncs\}@uni-heidelberg.de}}
\maketitle

\begin{abstract}
Class-agnostic counting methods enumerate objects of an arbitrary class, providing tremendous utility in many fields. Prior works have limited usefulness as they require either a set of examples of the type to be counted or that the query image contains only a single type of object. A significant factor in these shortcomings is the lack of a dataset to properly address counting in settings with more than one kind of object present. To address these issues, we propose the first Multi-class, Class-Agnostic Counting dataset (MCAC) and A Blind Counter (ABC123), a method that can count multiple types of objects simultaneously without using examples of type during training or inference. ABC123 introduces a new paradigm where instead of requiring exemplars to guide the enumeration, examples are found after the counting stage to help a user understand the generated outputs. We show that ABC123 outperforms contemporary methods on MCAC without needing human in-the-loop annotations. We also show that this performance transfers to FSC-147, the standard class-agnostic counting dataset. MCAC is available at \url{MCAC.active.vision} and ABC123 is available at \url{ABC123.active.vision}
\end{abstract}

\section{Introduction}
\label{Introduction}
Given an image and told to `count', a person would generally understand the intended task and complete it with accuracy even if there are multiple previously unseen classes of object present.
This natural human ability to count arbitrarily has not been modelled by today's methods. 
Most automated counting methods are class-specific \cite{cao2018scale, go2021fine}, counting objects of classes that were present during training. 
These methods are not generalisable and require retraining for each new type of object.
\textit{Class-agnostic} methods \cite{ranjan2021Famnet, shi2022represent} can count objects of an arbitrary type removing the need for retraining. 
However, they usually require an exemplar image as a prior on the class to count and only count a single class at a time.
Recently, \textit{exemplar-free} or \textit{zero-shot} class-agnostic counting methods \cite{ranjan2022exemplar, hobley2023LTCA} have been developed that do away with the need for exemplars to define type removing the need for human intervention during deployment. 
These methods either perform poorly \cite{ranjan2022exemplar} or are bounded to images that only contain a single class of object \cite{hobley2023LTCA, Djukic2023LOCA}.

We propose ABC123, a transformer-based, multi-class class-agnostic counter which does not need exemplars during training or inference.
ABC123 achieves this by first regressing density maps for each type present then enumerating the instances using integration.
As it is sometimes difficult to interpret what has been counted given only a density map and count, we design an example discovery stage which locates instances of the counted object.

A significant factor in the limitations of the current methods is the lack of a dataset for class-agnostic counting that includes images with more than one class present, as currently they all focus on the single-class scenario.
In order to train and evaluate our method, as well as other methods in multi-class settings we introduce MCAC, a new synthetic multi-class class-agnostic counting dataset.
We show that methods previously assumed to work in multi-class settings perform poorly on MCAC and that ABC123 significantly outperforms them while also generalising to other datasets. 

\begin{figure}[t]
\centering
\begin{minipage}{.46\textwidth}
  \centering
\includegraphics[width=\linewidth]{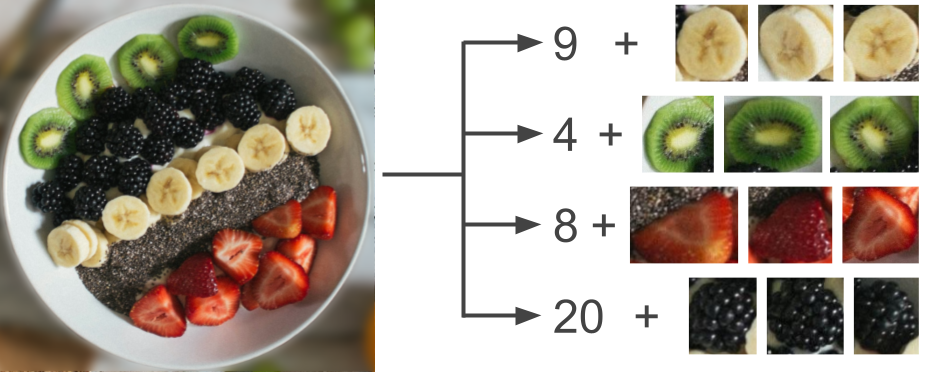}
  \captionof{figure}{\textbf{ABC123 counts objects of multiple unseen types}. Not only does our method not need exemplars to define the type to count, it finds examples of each type it has counted.}
\label{fig:teaser}
\end{minipage}%
\hspace{3mm}
\begin{minipage}{.5\textwidth}
  \centering
    \input{figures/blender_example_images}
    \captionof{figure}{\textbf{MCAC Contains images with up to 4 classes and up to 300 instances per class}. All objects have associated instance labels, class labels, bounding boxes, centre points, and occlusion percentages.}
  \label{dataset_teaser}
\end{minipage}
\end{figure}

\noindent Our main contributions are:
\begin{itemize}
    \item  We propose ABC123, the first exemplar-free multi-class class-agnostic counter and show it tackles multi-class counting effectively.
    \item We introduced MCAC, the first multi-class class-agnostic counting dataset and use it to demonstrate prior methods do not perform as expected in multi-class settings.
    \item We introduce the idea of \textit{example finding} to exemplar-free counting and demonstrate its utility in aiding a user in understanding what has been counted.
\end{itemize}

\noindent
In the remainder of this paper, we outline the relevant prior work in Section \ref{relatedwork} before introducing MCAC in Section \ref{sec:MCAC}. In Section \ref{sec:abc123}, we introduce ABC and detail our experimental setup. Section \ref{sec:results} presents our state-of-the-art results in single- and multi-class class-agnostic settings. Finally, we summarise the impact of the work as a whole.

\section{Related Work}
\label{relatedwork}
Class-specific counting methods 
aim to enumerate the instances of a single or small set of known classes \cite{cao2018scale, go2021fine, hoekendijk2021counting, xie2018microscopy}.
These methods struggle to adapt to novel classes, needing specific data and training for each type of object.
To address these issues, Lu et al.\ \cite{Lu18} proposed class-agnostic counting, a framework where inference-time classes are not present during training.
Still, most class-agnostic methods \cite{ranjan2021Famnet, yang2021cfoc,sokhandan2020few, shi2022represent}, require exemplar images of the test-time class. 
These methods generally work by creating a sufficiently general feature space and applying some form of matching to regress a density map of the counted objects.

Recent works, RepRPN \cite{ranjan2022exemplar}, CounTR \cite{liu2022countr}, ZSC \cite{xu2023zero}, CLIP-count \cite{jiang2023clip}, RCC \cite{hobley2023LTCA} and LOCA \cite{Djukic2023LOCA} do away with exemplar images at inference-time,
removing the need for intervention during deployment.
RepRPN is a two-step method which proposes regions likely to contain an object of interest and then uses them for an exemplar-based density map regression method. 
It proposes more than one bounding box and enumerates them separately. RepRPN performs poorly in comparison to other contemporary methods.
ZSC \cite{xu2023zero} uses a multi-stage process in which a text input is used to generate a generic image of the type to be counted that is then used to find exemplar patches. These exemplar patches then act as the input to an exemplar-based method \cite{shi2022represent}.
CounTR \cite{liu2022countr} uses a large vision-transformer encoder-decoder to regress a density map of instance locations. It is trained in a mixed few/zero-shot way, applying understanding gained from exemplar-based examples to exemplar-free cases. 
LOCA \cite{Djukic2023LOCA} also uses a vision transformer backbone and can perform both few- and zero-shot counting.
LOCA separately extract the shape and appearance of exemplar and non-exemplar objects to create a more informed object prototype. This prototype is then matched to areas of the image to generate a density map prediction.

It has been assumed that current exemplar-based methods can function in multi-class settings. However, this has not been proven rigorously as the main dataset for class-agnostic counting (FSC-133/147 \cite{ranjan2021Famnet, hobley2023LTCA}) contains only one labelled class per image. 
In fact, we show in \cref{sec:results} that these methods perform poorly in contexts with multiple types present.
FSC-133/147 being single-class has also explicitly motivated work such as RCC, which regresses a single scalar count from an image. It is trained without exemplar images and uses scalar supervision instead of density maps. 
Even with the constraint of only counting one kind of object and with no further direction on the type to count, RCC achieves competitive results with exemplar-based methods on FSC-147, showing the limitations of this dataset.

While large models with image inputs \cite{openai2023gpt4} like SAM \cite{kirillov2023segany} would seem to be able to effectively count objects of arbitrary types, in fact these methods have poor numerical understanding \cite{ma2023can}, and perform unsatisfactorily on counting tasks especially when images have small objects or a high density of objects.
Recently, Paiss \etal~\cite{paiss2023countclip} attempted to utilise multi-modal deep learning to count. Specifically, they introduce a method that teaches a CLIP model a coherent understanding of words related to counting. However, they are only able to achieve this up to 10 objects.

\section{MCAC Dataset}\label{sec:MCAC}

There are currently no datasets suitable for class class-agnostic counting problems with multiple types of object present at once. This significantly impacts the research into addressing these tasks. 
In order to facilitate the development of multi-class class-agnostic counting methods as well as the evaluation of prior work, we introduce MCAC, the first multi-class class-agnostic counting dataset.

While the deployment query scenario, `counting given an unlabelled image of objects', is natural, the training and quantitative evaluation of methods to address it is not.
To facilitate training and evaluation of methods in multi-class settings, we need images with multiple objects of multiple types. 
To evaluate a methods generalisability to unseen object types, the classes present in the images need to be mutually exclusive between training, validation and testing.
It is infeasible to gather natural images with (a) a wide variety of classes, (b) a wide variety of the number of times an object appears in an image, and (c) no repetition of the types of object between the train, test, and validation splits. Using synthetic images allows the above constraints to be satisfied while also providing a high level of precision and accuracy in the labels for each image. 
As shown in \cref{sec:results_fsc}, the understanding gained from training on synthetic data is general enough to apply to standard photographic datasets.

MCAC contains images with between 1 and 4 classes of object (mean of 1.75) and between 1 and 300 instances per class (mean of 47.66). 
MCAC has three data splits: training with 4756 images drawn from 287 classes; validation 2413 images drawn from 37 classes, and testing with 2114 images drawn from 19 classes.
MCAC-M1 is the single-class subset of the MCAC images which have only one class present per image. MCAC-M1 totals 4259 images, with a mean of 114.89 instances per image. These distributions were designed to replicate that of real-world counting tasks.

All instances in an image have associated class labels, model labels, center coordinates, bounding box coordinates, segmentation maps, unoccluded segmentation maps, and occlusion percentages. The occlusion percentage is calculated as $1 - \frac{A_0}{A_1}$, where $A_0$ is the number of pixels in the final image and $A_1$ is the number of pixels that would be seen if the object was unoccluded and completely within the bounds of the image.

Objects are `dropped' into the scene, ensuring random locations and orientations.
As objects in real settings often vary in size, we vary the size of objects by $\pm$50\% from a random nominal size.
We also vary the number, location, and intensity of lights present.
Models and textures are drawn from ShapeNetSem \cite{savva2015semgeo}.

Both exemplar-based and exemplar-free methods bump into problems of ambiguity. If there are objects of varied levels of generality, which boundary should be used? 
For example, on a chess board with a single white pawn as the exemplar, should the count be of all the pieces, all the white pieces, all the white pawns, all the pawns, and so on? 
\Cref{fig:blender_fsc_examples} shows examples from FSC-147 of cases with an ambiguity of what is to be counted.

Given the infeasibility of defining every possible way of grouping the objects present in an image, we define a single way of grouping the objects: an identical mesh and texture, independent of size or orientation. 
We do, however, acknowledge the existence of other 
\textit{valid-but-unknown} counts, the unlabelled ways of grouping the objects. 

\begin{figure}[t]
    \centering
    \includegraphics[width=0.98\linewidth]{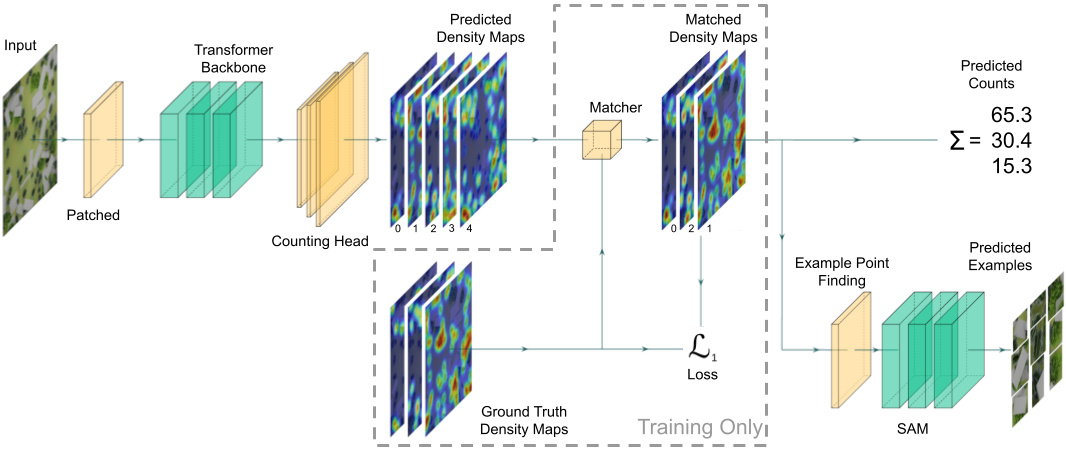}
        \caption{
    \textbf{The ABC123 pipeline.} 
    Our method learns to count objects of multiple novel classes without needing exemplar images. During training and quantitative evaluation, the matcher aligns the unguided predictions to the ground truth labels. To aid a user in understanding the results, the example prediction stage locates instances associated with each generated count.
    }
    \label{pipeline}
\end{figure}

\section{Method}\label{sec:abc123}
Our method, ABC123, takes an image with multiple instances of objects of multiple types and regresses the count of each type. This is achieved \textit{blind}, i.e.\ on objects of arbitrary classes with no requirement to have seen the object class during training or to have an exemplar image to define the type during inference.
We achieve this by first regressing density maps for each type then enumerating the instances using integration. To facilitate training and evaluating ABC123 in an exemplar-free way, we propose a matching stage.
To further increase the interpretability of the outputs of ABC123, we design an example discovery stage which finds specific instances of the counted object.
The pipeline of ABC123 is presented in \cref{pipeline}.

\subsection{Density Map Regression}
For each image, there are $m$ classes present, each with an associated ground truth count $y$ and density map $d$. We regress $\hat{m}$ counts and density map predictions, $\hat{y}$ and $\hat{d}$ respectively. $\hat{m}$ acts as upper-bound of the number of counts ABC123 can regress.

\begin{equation}
    \hat{y} = \sum_{h,w}{\hat{d}_{(h,w)}}
\end{equation} 
where $\hat{d}_{(h,w)}$ denotes the density value for pixel (h, w).
We achieve this by using $\hat{m}$ convolutional up-sampling heads on top of a vision transformer backbone \cite{dosovitskiy2020vit}. 
We use a vision transformer backbone due to its globally receptive field and self-attention mechanism, which Hobley and Prisacariu \cite{hobley2023LTCA} showed is helpful to generating a complex understanding in counting settings.
Each head regresses a single pixel-wise density map prediction and count prediction from a patch-wise low-resolution high-dimensional feature space.
Similar to other contemporary methods \cite{ranjan2021Famnet, liu2022countr, yang2021cfoc, lin2021object}, we use the pixel-wise error $||d - \hat{d}||_1$ as our loss, where $d$ and $\hat{d}$ are the ground truth and predicted density maps.

\subsection{Matching}
In single-class or exemplar-based settings, there is a single prediction-label pair. 
However, in multi-class exemplar-free settings, there are multiple predictions as well as multiple labels, without a clearly defined pairing.
This resembles other open-set problems like class-discovery \cite{troisemaine2023novel, 5454426} and clustering \cite{hobley2023dms,pmlr-v48-xieb16}, where the number and cardinality of new classes is not necessarily known. 
In keeping with these fields and to facilitate training and quantitative evaluation, we find correspondences between the set of $m$ known counts and the set of $\hat{m}$ predicted counts. 
The correspondence matrix is defined as $\mathcal{X} = \{0,1\}^{\hat{m}\times m}$ where  $\mathcal{X}_{i,j}=1$ iff prediction $i$ is assigned to label $j$.
A problem instance is described by an $\hat{m}\times m$ cost matrix $\mathcal{C}$, where $\mathcal{C}_{i,j}$ is the cost of matching prediction $i$ and ground truth label $j$. The goal is to find the complete assignment of predictions to labels of minimal cost.
Formally, the optimal assignment has cost
\begin{equation}\label{eq:Bipartite_matching}  \min_{\mathcal{X}}\sum_{i=0}^{m}\sum_{j=0}^{\hat{m}}\mathcal{C}_{i,j} \cdot \mathcal{X}_{i,j}
\end{equation}

Specifically, our cost function is defined as the the pixel-wise distance of the normalised ground truth density map $d_i$ and the predicted density map $\hat{d}_j$. 

\begin{equation} \label{eq:MAE_cost}
\mathcal{C}_{i,j} = \left|\left| \frac{d_i}{||d_i||_2} -  \frac{\hat{d}_j}{||\hat{d}_j||_2} \right|\right|_2 
\end{equation}

The normalisation ensures the matching is done on the locality of the counted objects rather than the magnitude of the prediction itself.
We use the Hungarian algorithm, specifically the Jonker-Volgenant algorithm outlined in Crouse~\cite{Crouse2016hungarianmatching}, to solve for $\mathcal{X}$ robustly.

The supervision loss for each image is the sum of the L$_1$ difference of  the ground truth density maps and their matched predictions as:
\begin{equation} \label{eq:total_loss}
\mathcal{L} = \sum_{i,j}^{m,\hat{m}}||d_i - \hat{d}_j||_1 \cdot \mathcal{X}_{i,j}
\end{equation}

It should be noted that every label has an associated prediction, but the inverse is not the case as generally $\hat{m} > m$.
This means we do not impose a loss on the unmatched density maps.
This allows the network to generate more nuanced count definitions as it does not punish valid-but-unknown counts which are likely present in any counting setting.
As is usual \cite{troisemaine2023novel, 5454426,pmlr-v48-xieb16}, we use the same matching procedure to evaluate our performance at inference-time. However, during deployment, when there are no ground-truth density maps, the matching is both unnecessary and impossible. We instead combine any similar predictions, remove predictions of zero, and present the user with the predictions.

\subsection{Example Discovery}
\begin{figure}[t]
    \input{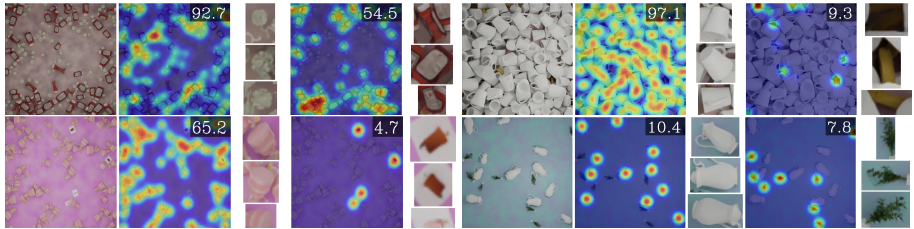}
    \caption{
    \textbf{Example Finding.} 
    Instead of using exemplars to define the count, we count `blind' and then find meaningful bounding boxes to aids a user in understanding what has been counted. The examples are found using the query image, learnt features, our regressed density maps and a SAM network.} 
        \label{fig:example_discovery}
\end{figure}

While exemplar-free counting saves a user time, as no manual intervention is required, it does require the user to interpret the results. A set of scalar counts or density maps can be unclear as it is not always obvious which count corresponds to which type of object in the input image, especially in high density situations; see \Cref{fig:example_discovery}.
To aid the user in understanding to which class a generated count corresponds, we propose flipping the usual exemplar-based paradigm.
Instead of using exemplar images to define the type to count, we find examples of the type that was counted.

To find examples corresponding to a given count, we first find example points which are high in the corresponding density map while low in the others. To increase the diversity of the found examples we select the points with the largest latent feature distance. We use these points as seed inputs for a pre-trained segmentation method, SAM \cite{kirillov2023segany}.
The end user is presented with cropped areas of the query image centred on these segmentations, as in  \cref{fig:example_discovery}.

\begin{figure}[t]
    \centering    
    \includegraphics[width=0.49\linewidth]{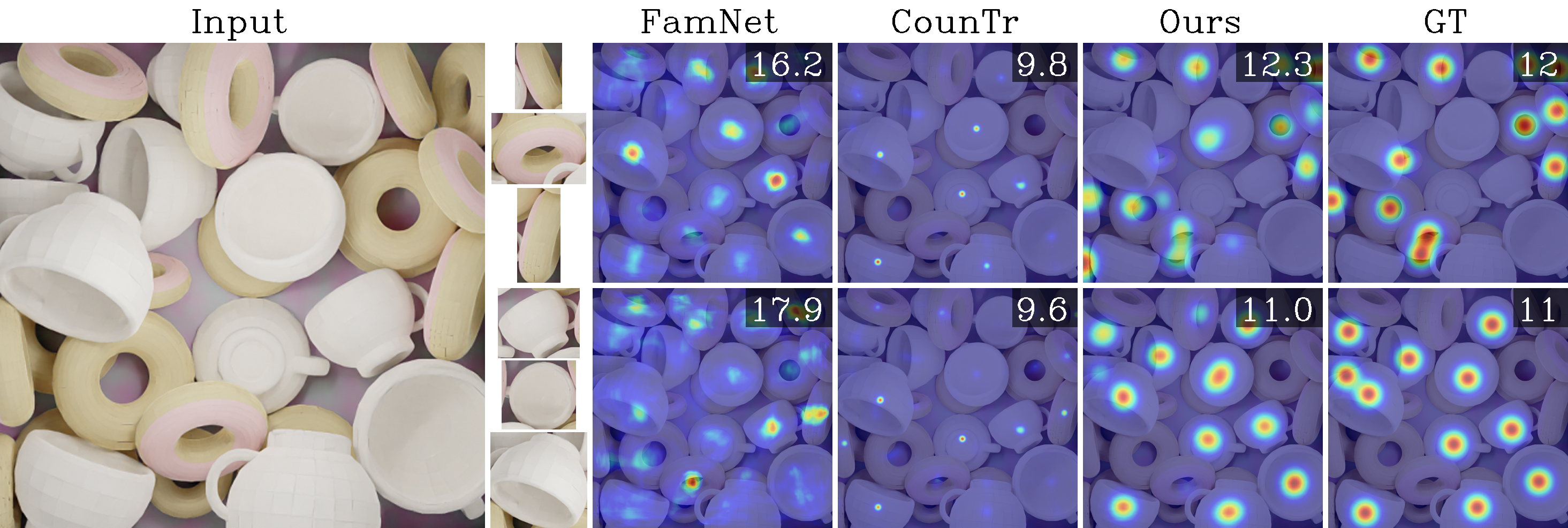}%
    \hspace{0.01\linewidth}
    \includegraphics[width=0.49\linewidth]{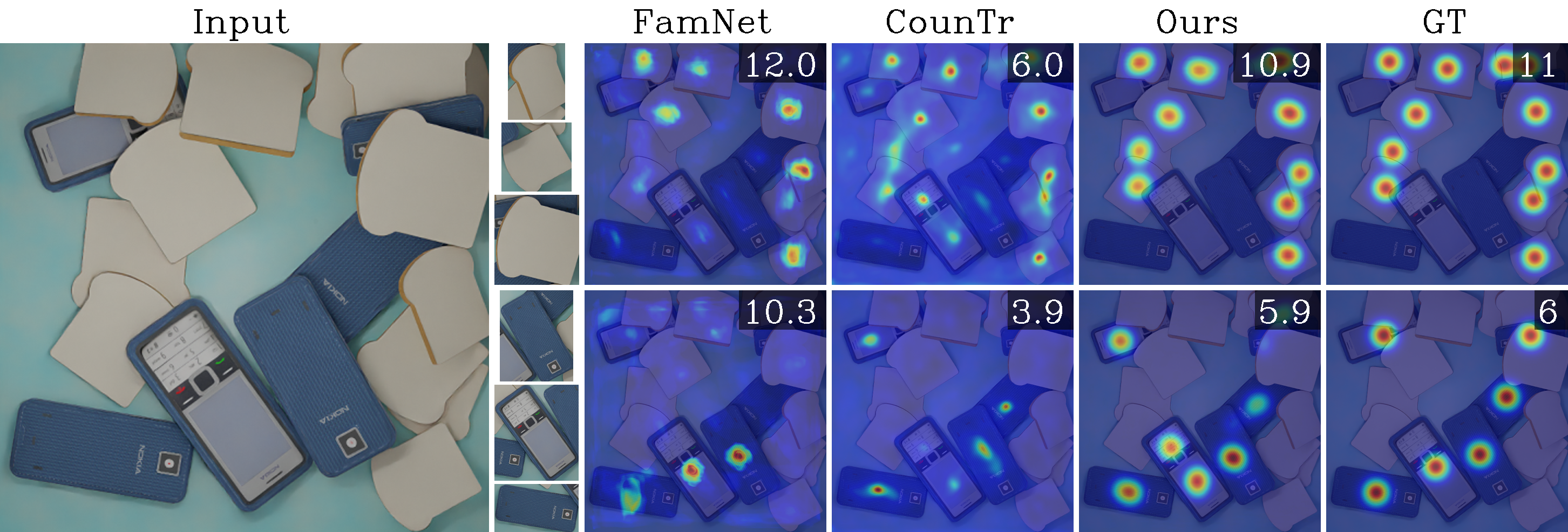}\\
    \vspace{0.5mm}%
    \includegraphics[width=0.49\linewidth]{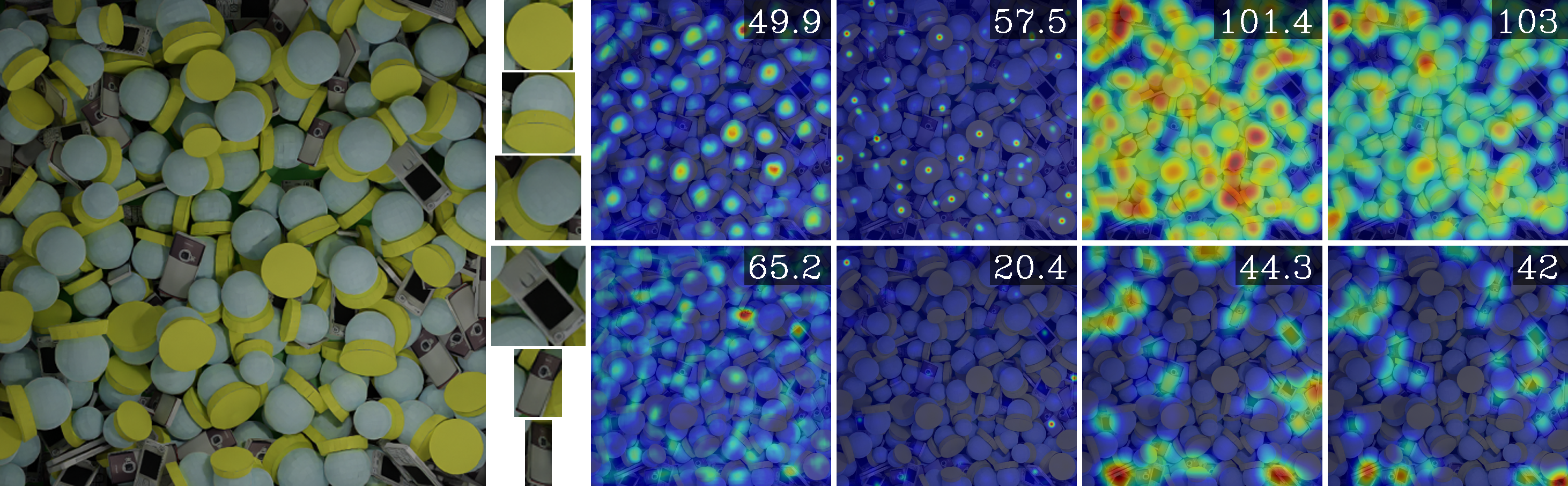}%
    \hspace{0.01\linewidth}
    \includegraphics[width=0.49\linewidth]{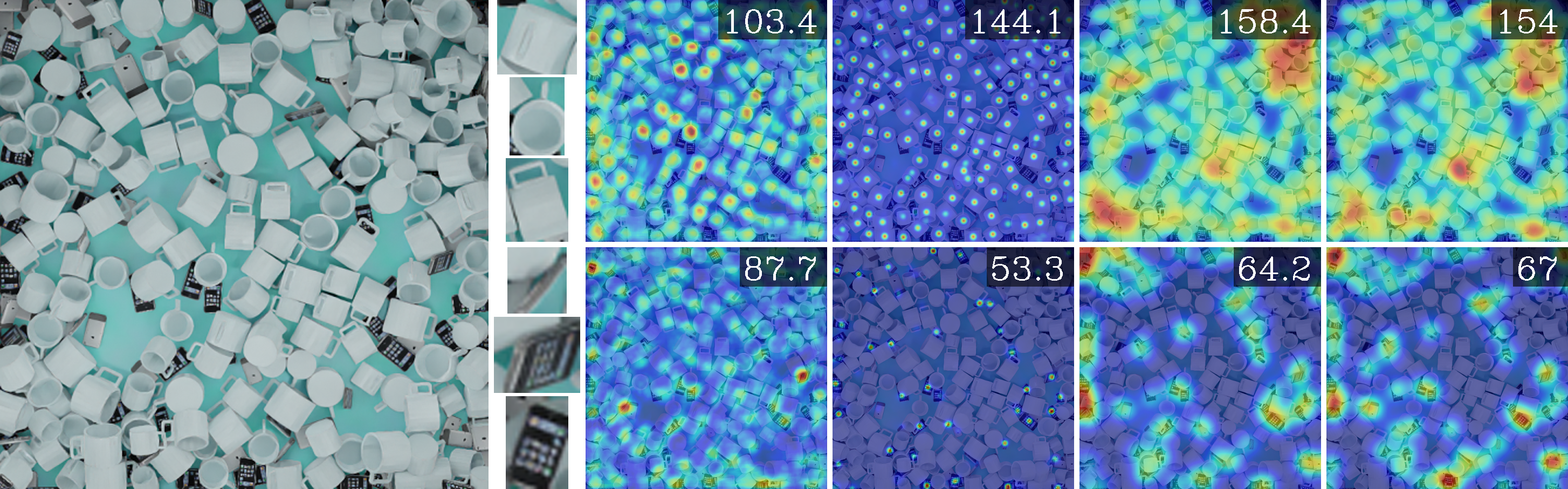}\\
        \caption{
    \textbf{Comparison to other methods on MCAC.} 
    ABC123 produces more accurate results than the exemplar-based methods without using exemplar images.
    The ground truth (GT) and  predicted counts are shown in the top right corner of their respective density maps.
    }
    \label{fig:MCAC_vis}
\end{figure}

\begin{figure}[t]
    \centering    
    \includegraphics[width=\linewidth]{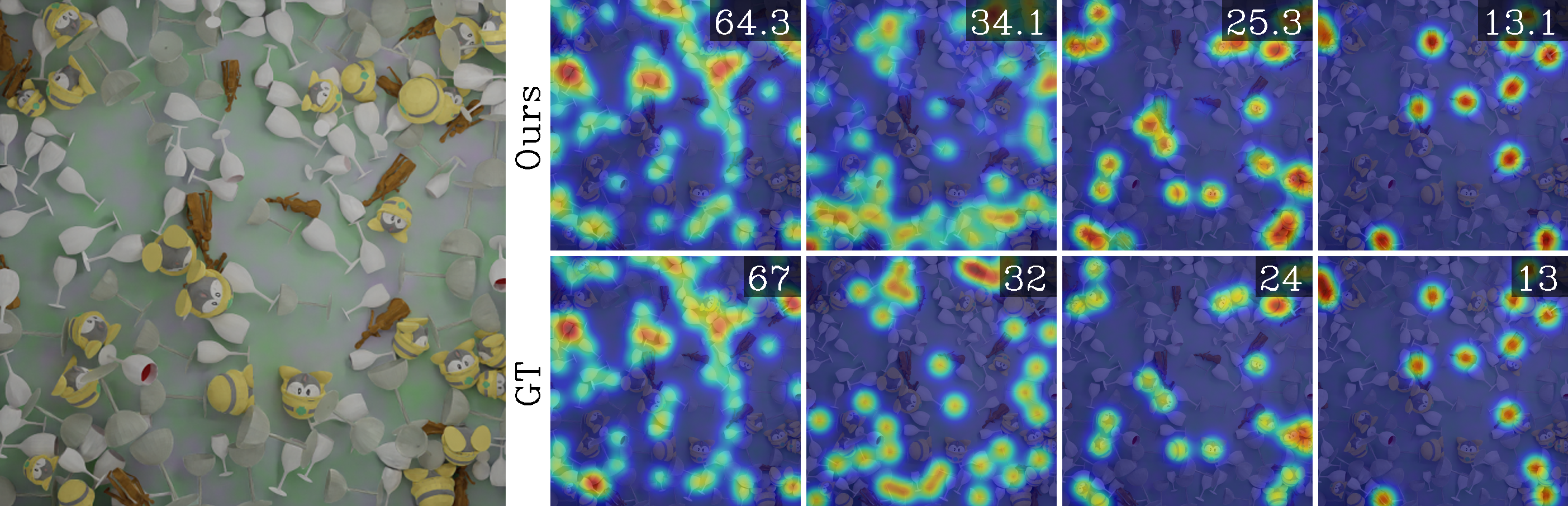}
    \includegraphics[width=\linewidth]{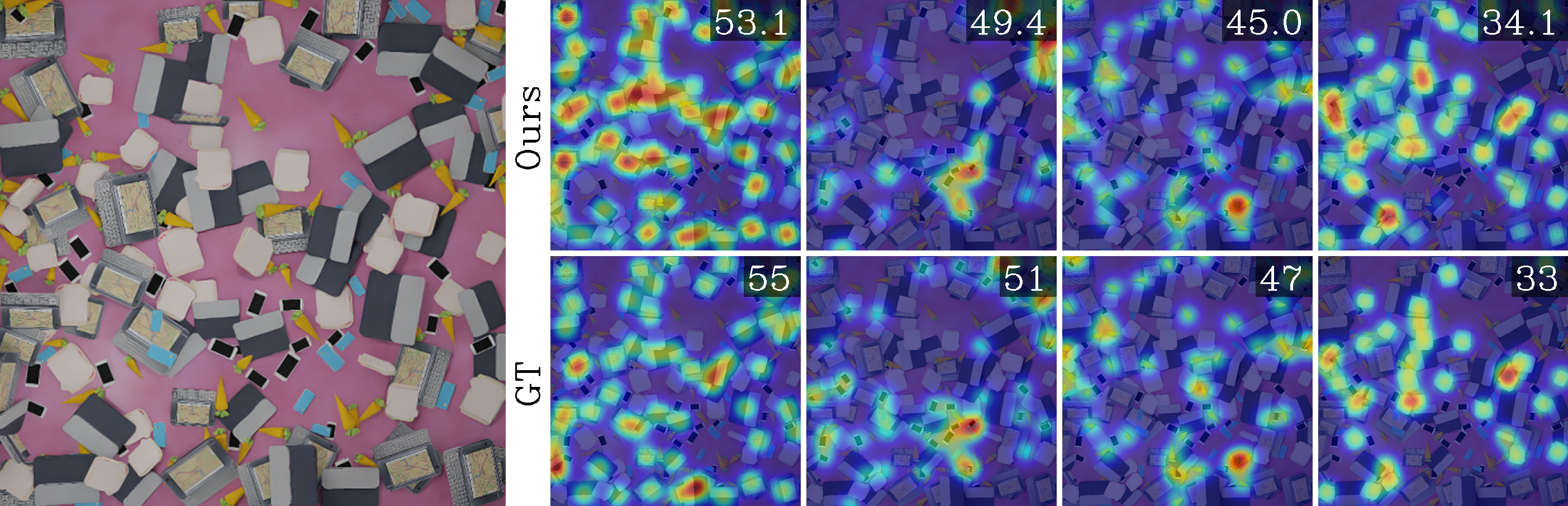} 
        \caption{
    \textbf{Results of our method on images with 4 classes.} 
    ABC123 is able to generate accurate counts and meaningful density maps from images with four novel classes. MCAC has between one and four classes of object per image.
    }
    \label{fig:MCAC_vis_4_classes}
\end{figure}

\subsection{Experiments}
We use a ViT-Small \cite{dosovitskiy2020vit} backbone due to its lightweight nature and for comparison to methods that use the ResNet-50 backbone, such as FamNet and BMNet \cite{ranjan2021Famnet, shi2022represent}.
ViT-S has a similar number of parameters (21M vs 23M), throughput (1237im/sec vs 1007im/sec), and supervised ImageNet performance (79.3\% vs 79.8\%) as ResNet-50 \cite{touvron2021training}.
Our choice of ViT-S for ABC123 limits the input resolution to (224$\times$224) as opposed to the ($\geq$384$\times$384) resolution used by the method we compare to with ResNet-50 or larger ViT backbones.
The effect of using ResNet or ViT-S backbones in various counting methods was discussed in detail in Hobley and Prisacariu \cite{hobley2023LTCA} where they found that neither was inherently superior and the performance varied dependant on the method.

Since vision transformers typically demand substantial training data, we initialise our transformer backbone with weights from Caron \etal~\cite{caron2021DINO}.
This self-supervised pre-training endows the network with an understanding of meaningful image features prior to exposure to our dataset and without supervision. This reduces the risk of overfitting when the model is then trained on MCAC.

Our counting heads, comprised of 3 Conv-ReLU-Upsample blocks, increase the patch-wise resolution of the trained counting features from $k \times (28\times28)$ to a pixel-wise density map prediction of $\hat{m} \times (224\times224)$,
where $k$ is the dimensionality of the transformer features and $\hat{m}$ is the number of predicted counts. For ABC123, $k = 384$.
We set $\hat{m} = 5$ to ensure there is the capacity to generate a count per defined class in MCAC and at least one valid-but-unknown count. 

ABC123 trains in less than eight hours using two 1080Tis. It takes less than two hours to train just the head with a frozen backbone (ABC123\textsmaller[2]{\SnowflakeChevron}).
During training we use an Adam optimiser, a batch-size of 2 and a learning rate of $3*10^{-5}$ which halves every 35 epochs for a total of 100 epochs.
The example discovery stage uses a frozen pretrained ViT-B SAM model \cite{kirillov2023segany}. 

Here we lay out our usage of MCAC, which we used to generate our results. We recommend future works use a similar approach and we will release code for a PyTorch dataset to enable easy adoption.
We exclude objects that are more than 70\% occluded by either other objects or the edge of the frame. 

MCAC enables the use of true pixel-wise density maps. 
While this increases the accuracy of our method, we found this significantly decreased the performance of other methods, especially those with test-time adaptations.
For fairness, we use the standard \cite{ranjan2021Famnet, liu2022countr, yang2021cfoc, lin2021object} pseudo-density-map for all methods, including our own. These are generated by placing a Gaussian kernel centred on the center pixel of each object.

When training exemplar-based methods, we take bounding boxes randomly from instances with less than 30\% occlusion.
We evaluate these methods using the bounding boxes of the three least occluded instances.

\section{Results}\label{sec:results}
We evaluate our method against two trivial baseline methods, predicting the training-set mean or median count for all inference images. As there are no previous multi-class exemplar-free class-agnostic counting methods, we compare ABC123 to exemplar-based methods using separate exemplars from each of the classes present.
We compare our method to FamNet \cite{ranjan2021Famnet}, BMNet \cite{shi2022represent} CounTR \cite{liu2022countr} and LOCA \cite{Djukic2023LOCA} on MCAC and FSC-133/147 as these are the current state-of-the-art methods with publicly available implementations.
Additionally, we also compare to RCC \cite{hobley2023LTCA} and CounTR in its zero-shot configuration on MCAC-M1, the subset of MCAC with only a single type of object present per image.

As in Xu \etal~\cite{xu2023zero}, we use Mean Absolute Error (MAE), Root Mean Squared Error (RMSE), Normalized Absolute Error (NAE), and Squared Relative Error (SRE) to evaluate the performance of each method.

{\small
\begin{equation}
    \text{MAE} = \frac{1}{nm} \sum_{j=1}^n\sum_{i=1}^{m_{j}} |y_i-\hat{y_i}|, \hspace{10mm} \text{RMSE} = \sqrt{\frac{1}{nm} \sum_{j=1}^n\sum_{i=1}^{m_{j}} (y_i-\hat{y_i})^2} 
\end{equation}
\begin{equation}
\text{NAE} = \frac{1}{nm} \sum_{j=1}^n\sum_{i=1}^{m_{j}} \frac{|y_i-\hat{y_i}|}{y_i}, \hspace{10mm} \text{SRE} = \sqrt{\frac{1}{nm} \sum_{j=1}^n\sum_{i=1}^{m_{j}} \frac{(y_i-\hat{y_i})^2}{y_i}}
\end{equation}
}

\noindent \looseness=-1 where $n$ is the number of test images, $m_j$ is the number of classes in image $j$, $y_i$ and $\hat{y_i}$ are the ground truth and predicted number of objects of class $i$ in image $j$.

\subsection{MCAC}
We achieve significantly better result compared to FamNet, BMNet, CounTR and LOCA on MCAC both quantitatively and qualitatively without needing exemplars; see \cref{blender_results_table} for results and \cref{fig:MCAC_vis} for comparative examples.
As seen in \cref{fig:MCAC_vis}, FamNet often fails to discriminate between objects of different classes when they are visually similar or in high density applications.
Both quantitatively and qualitatively, BMNet and CounTR outperform FamNet. However, in many cases, they appear to count the `most obvious' objects in the image regardless of the provided exemplar images.
This behaviour is present but is less prevalent with LOCA.
Our method performs well on images with up to 4 classes even when they have high intra-class appearance variation, such as having different colours on different sides, and low inter-class variation; see \cref{fig:MCAC_vis_4_classes}.
A downside to current exemplar-based class-agnostic counting methods is that while they have some multi-class capabilities, they all take a single exemplar at a time and produce only one count. This is slow and inefficient as compared to our approach which generates all counts simultaneously.

As would be expected, the performance of all methods improves when evaluating on MCAC-M1, the images from MCAC with only a single class present; see \cref{blender_single_class_results}. This is due to a lack of ambiguity of the type to be counted. This was more significant when the methods were trained on MCAC-M1 instead of MCAC. In this training configuration, the methods generally learnt a broader definition of similarity as there was no chance they would accidentally combine classes or count instances from another class.
RCC performs well on MCAC-M1, showing the strength of the simple count-wise loss in cases where there is little ambiguity as to what is to be counted.
In contrast to other methods, ABC123 trained on MCAC-M1 has similar performance to when it is trained on the full MCAC dataset, demonstrating that it avoids issues concerning intra-class variance and combining classes. Training ABC123 with only a single head ($\hat{m}=1$) and no matching stage has very similar performance to using its default ($\hat{m}=5$) configuration with a matching stage. This increases our belief that the matching head does not provide an unfair advantage to our method's quantitative results.

\begin{table}[t]
    \centering
    \caption{\textbf{Comparison to SOTA methods on MCAC.} We significantly outperform methods which use exemplar images and test-time adaptation without requiring them. ABC123\textsmaller[2]{\SnowflakeChevron} denotes our method trained with a frozen pre-trained backbone.
    }
    \fontsize{9}{9}\selectfont
        
\begin{tabular}{lcrrrrrrrr}
 \toprule
%  \multicolumn{5}{c}{\textbf{FSC-147}}\\
%  \midrule
 & & \multicolumn{4}{c}{Val Set} & \multicolumn{4}{c}{Test Set} \\
 \cmidrule(r){3-6} \cmidrule(r){7-10}

\multicolumn{1}{c}{Method} & \multicolumn{1}{c}{Shots} & \multicolumn{1}{c}{MAE} & \multicolumn{1}{c}{RMSE} & \multicolumn{1}{c}{NAE} & \multicolumn{1}{c}{SRE} & \multicolumn{1}{c}{MAE} & \multicolumn{1}{c}{RMSE} & \multicolumn{1}{c}{NAE} & \multicolumn{1}{c}{SRE} \\
\midrule
Mean & N/A  &  39.87 & 53.56 & 3.07 & 11.40 & 42.67 & 59.68 & 2.79 & 10.93 \\
Median & N/A  & 36.25 & 58.15 & 1.51 & 6.70 & 39.81 & 65.36 & 1.38 & 6.73\\
\midrule
\textit{Exemplar-based} & & & & \\
FamNet+ \cite{ranjan2021Famnet} & 3 & 24.76 & 41.12 & 1.12 &  6.86 &  26.40 & 45.52 & 1.04 &  6.87 \\ 
BMNet+ \cite{shi2022represent} & 3 &   15.83 &  27.07 &  0.71 & 4.97  & 17.29 & 29.83 & 0.75 & 6.08   \\ 
CounTR \cite{liu2022countr}  & 3 & 15.07 & 26.26 & 0.63 & 4.79 & 16.12 & 29.28 & 0.67 & 5.71 \\
LOCA \cite{Djukic2023LOCA}  & 3 &  10.45 & 20.81 & 0.43 & 4.18 &   10.91 & 22.04 & 0.37 & 4.05 \\
\midrule
\textit{Exemplar-free} & & & & \\
ABC123 \textsmaller[2]{\SnowflakeChevron} & 0 & 14.64 & 23.67 & 0.46 & 2.97 & 15.76 & 25.72 & 0.45 & 3.11 \\
ABC123 & 0 & \textbf{8.96} & \textbf{15.93} & \textbf{0.29} & \textbf{2.02} & \textbf{9.52} & \textbf{17.64} & \textbf{0.28} & \textbf{2.23}  \\

\bottomrule
\end{tabular}

    \label{blender_results_table}
\end{table}

\begin{table}[t]
    \centering
    \caption{\textbf{Comparison to SOTA methods on MCAC-M1.} MCAC-M1 is the subset of MCAC with only one class present per image. Methods are either trained on the full multi-class dataset (\cmark) or MCAC-M1 (\xmark). `$\hat{m}$' denotes the number of predictions the method generates per query and `Shots' denotes the number of exemplar images per query at inference time. CounTR\dag \ is an exemplar-free adaption of CounTR as in Hobley and Prisacariu \cite{hobley2023LTCA}.  ABC123 outperforms other methods when trained in single or multi-class settings.
    }
    \resizebox{\linewidth}{!}{%
    \begin{tabular}{lcccrrrrrrrr}
 \toprule
 & & & & \multicolumn{4}{c}{Val Set} & \multicolumn{4}{c}{Test Set} \\
 \cmidrule(r){5-8} \cmidrule(r){9-12}

\multicolumn{1}{c}{Method} & \multicolumn{1}{c}{Multi-Class Training} & \multicolumn{1}{c}{Shots} & \multicolumn{1}{c}{$\hat{m}$} & \multicolumn{1}{c}{MAE} & \multicolumn{1}{c}{RMSE} & \multicolumn{1}{c}{NAE} & \multicolumn{1}{c}{SRE} & \multicolumn{1}{c}{MAE} & \multicolumn{1}{c}{RMSE} & \multicolumn{1}{c}{NAE} & \multicolumn{1}{c}{SRE} \\

\midrule
Mean & N/A & N/A & N/A & 
53.36 & 67.14 & 3.53 & 13.46 & 58.54 & 75.58 & 3.37 & 13.27\\
Median & N/A & N/A & N/A &
45.98 & 76.64 & 1.08 & 6.68 &
51.35 & 86.61 & 1.03 & 7.00\\

\midrule
\multicolumn{3}{l}{\textit{Exemplar-based Training}}& & \\
FamNet+ \cite{ranjan2021Famnet} & \cmark & 3 & 1 & 24.97 & 48.63 & 0.36 & 3.79 & 28.31 & 54.88 & 0.35 & 3.97 \\ 
FamNet+ \cite{ranjan2021Famnet} & \xmark & 3 & 1 & 12.54 & 24.69 & 0.37 & 4.71 & 13.97 & 26.19 & 0.25 & 2.12 \\
BMNet+ \cite{shi2022represent} & \cmark & 3 & 1 & 11.70 & 23.08 & 0.26 & 2.39 & 11.57 & 22.25 & 0.24 & 1.96 \\ 
BMNet+ \cite{shi2022represent} & \xmark & 3 & 1 & 6.82 & 12.84 & 0.25 & 2.95 & 8.05 & 14.57 & 0.19 & 1.43 \\ 
CounTR \cite{liu2022countr} & \cmark & 3 & 1 & 11.44 & 21.37 & 0.33 & 2.36 & 10.91 & 21.70 & 0.29 & 2.01 \\
CounTR \cite{liu2022countr} & \cmark & 0 & 1 & 13.57 & 25.53 & 0.30 & 2.48 & 13.09 & 25.72 & 0.29 & 2.41 \\
CounTR \cite{liu2022countr} & \xmark & 3 & 1 & 9.00 & 16.91 & 0.41 & 3.56 & 9.96 & 18.92 & 0.38 & 2.93 \\
CounTR \cite{liu2022countr} & \xmark & 0 & 1 & 9.16 & 17.13 & 0.42 & 3.56 & 10.10 & 19.10 & 0.40 & 3.02 \\

LOCA \cite{Djukic2023LOCA} & \cmark & 3 & 1 & 5.62 & 12.24 & \textbf{0.15} & \textbf{1.73} & \textbf{6.25} & \textbf{13.09} & \textbf{0.12} & \textbf{1.14} \\
LOCA \cite{Djukic2023LOCA} & \xmark & 3 & 1 & \textbf{5.01} & \textbf{11.47} & 0.22 & 3.35 & 6.52  & 13.37  & 0.15  & 1.36

 \\

\midrule
\multicolumn{3}{l}{\textit{Exemplar-free Training}}& & \\
CounTR\dag \cite{liu2022countr} & \xmark & 0 & 1 & 11.46 & 21.24 & 0.35 & 2.78 & 12.54 & 23.84 & 0.31 & 2.38 \\
RCC \cite{hobley2023LTCA} & \xmark & 0 & 1 & 7.78 & 15.40 & 0.24 & 2.71 & 8.81 & 16.92 & 0.19 & 1.73 \\
LOCA \cite{Djukic2023LOCA} & \xmark & 0 & 1 & \textbf{5.46} & \textbf{11.74} & 0.22 & 2.90 & 6.94 & \textbf{14.58} & 0.19 & 1.70 \\

ABC123 \textsmaller[2]{\SnowflakeChevron} & \xmark & 0 & 5 & 10.78 & 18.83 & 0.28 & 1.97 & 13.23 & 24.57 & 0.29 & 2.39 \\
ABC123 \textsmaller[2]{\SnowflakeChevron} & \xmark & 0 & 1 & 11.38 & 19.73 & 0.40 & 3.51 & 14.31 & 25.40 & 0.37 & 2.79 \\
ABC123 \textsmaller[2]{\SnowflakeChevron} & \cmark & 0 & 5 & 10.98 & 18.85 & 0.30 & 1.93 & 13.13 & 23.93 & 0.29 & 2.18 \\
ABC123 & \xmark & 0 & 5 & 5.82 & \textbf{11.74} & \textbf{ 0.15} & \textbf{1.22} & 7.54 & 15.30 &0.21 & 1.87 \\
ABC123 & \xmark & 0 & 1 & 5.85 & 12.91 &0.24 & 3.37 & 7.53 & 15.69 & 0.22 & 2.19 \\
ABC123 & \cmark & 0 & 5 & 6.08 & 12.62 & 0.16 & \textbf{1.22} & \textbf{6.82} & 14.70 & \textbf{0.16} & \textbf{1.51} \\
\bottomrule
\end{tabular}

    }
    \label{blender_single_class_results}
\end{table}
\subsection{Applicability to FSC-147/133}\label{sec:results_fsc}
\begin{figure}[t]
\centering
    \input{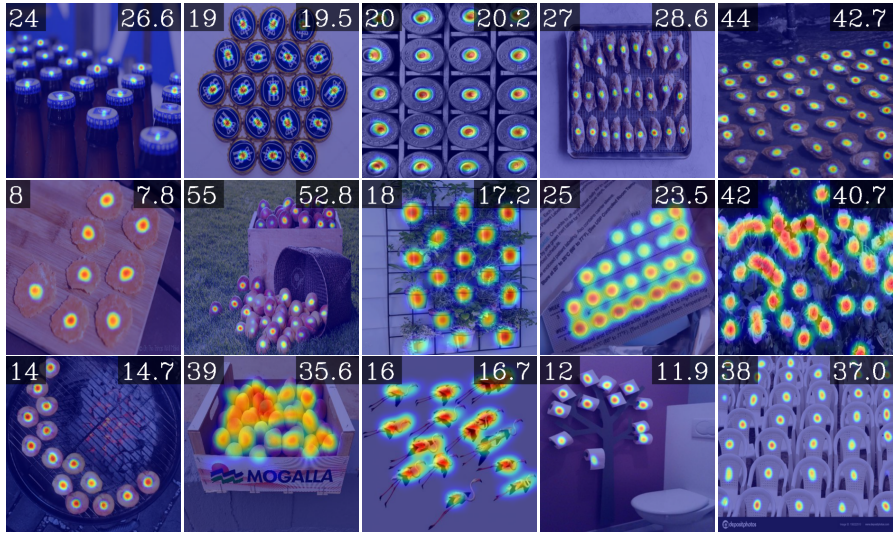}
    \caption{
     \textbf{ABC123 trained on MCAC applied to FSC-147 produces accurate counts.}    
    The ground truth and predicted counts are in the top left and top right corners.
    }
    \label{fig:FSC_easy}
\end{figure}

ABC123 trained on only MCAC, a synthetic dataset, produces accurate results and outperforms other contemporary methods when applied to FSC-133/147, the standard, more complex, photographic dataset,  as seen in \Cref{fig:FSC_easy} and \Cref{tab:fsc147_numerical_table}. 
As the standard benchmark evaluation metrics rely on absolute count error they are all very sensitive to even a small number of very dense images.
This phenomenon was discussed in Hobley and Prisacariu \cite{hobley2023LTCA}. 
We found that the errors on the few images with counts between 300 (the largest count in MCAC) and 3000 (the largest count in FSC) corrupted the metrics, making comparison to other literature more difficult. 
For this reason these images are excluded from the quantitative evaluation in \Cref{tab:fsc147_numerical_table}. These exclusions amount to 3.0\% of the validation set and 1.1\% of the test set. It should be noted that the relative rankings of the methods with and without this exclusion remain the same. 

\begin{figure}[t]
    \centering
    \includegraphics[width=0.99\linewidth]{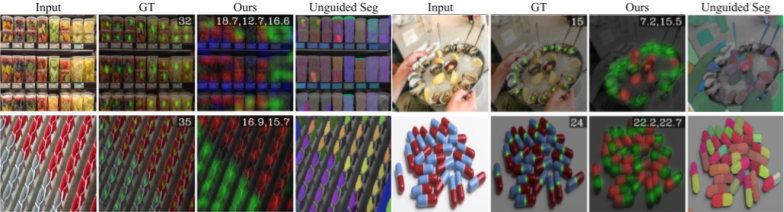}
    \caption{
    \textbf{ABC123 trained on MCAC applied to Ambiguous Images in FSC-147}. 
    In cases with class ambiguity, ABC123 often finds valid-but-unknown counts, i.e.\ they don't align with the human annotations. Similar to an unguided segmentation method (right), ABC123 discovers unlabelled classes, divides labelled classes into sub classes, or counts component parts of objects.
    The unguided segmentations are coloured by latent feature to demonstrate how they would be grouped.
    }
    \label{fig:blender_fsc_examples}
\end{figure}

While ABC123 performs well on FSC, it often finds valid-but-unknown counts. As seen in \cref{fig:blender_fsc_examples}, the generated counts are correct for the type of object counted, but the type counted may not be aligned with the labels in the original dataset. Classes are often divided into sub-classes, and unlabelled classes are discovered.

This is due to a difference in definition of what is similar. MCAC associates a count with objects of the same mesh and texture, however, FSC is labelled by hand uses high-level semantic understanding so often groups objects with significantly different geometries, colours, or textures. 
Interestingly, an unguided segmentation method \cite{kirillov2023segany}, which identifies instances' relations often finds the same class divisions as ABC123, shown in \Cref{fig:blender_fsc_examples}.
To generate quantitative results, we borrow the approach of other open set methods \cite{ma2023can, hobley2023dms, kirillov2023segany}, combining sub-classes. 
We perform this mapping by combining the density maps of sub-class counts, either by the summation of both the separate counts or by trying to combine the density maps using the maximum density at a given point and then counting the instances. The maximum density map configuration produces results which are competitive with other contemporary methods while the density map summation is clearly SOTA, as presented in \Cref{tab:fsc147_numerical_table}. The two approaches only differ in cases where sub-class density maps overlap. 

\begin{table}[t]
    \centering
     \caption{\textbf{Comparison to SOTA methods when trained on MCAC and applied to the cases in FSC147 with fewer than 300 objects}. 
     Combining sub-class density maps with a sum rather than a max is more effective as it is more accurate in cases where instances of the sub classes are spatially close or overlapping as both instances are counted completely.
    CounTR \dag \enspace is an exemplar-free modification of CounTR.
    }
    \fontsize{9}{9}\selectfont
    \begin{tabular}{lccrrrrrrrr}
 \toprule
 & & \multirow{2}{*}{\shortstack{\vspace{0.6mm}\\Sub-Class \hspace{1mm}\\ Combine}} & \multicolumn{4}{c}{Val Set} & \multicolumn{4}{c}{Test Set} \\
 \cmidrule(r){4-7} \cmidrule(r){8-11}

\multicolumn{1}{c}{Method} & \multicolumn{1}{c}{Shots} &  & \multicolumn{1}{c}{MAE} & \multicolumn{1}{c}{RMSE} & \multicolumn{1}{c}{NAE} & \multicolumn{1}{c}{SRE} & \multicolumn{1}{c}{MAE} & \multicolumn{1}{c}{RMSE} & \multicolumn{1}{c}{NAE} & \multicolumn{1}{c}{SRE} \\
\midrule
\multicolumn{5}{l}{\textit{Exemplar-based}}\\ 
FamNet+ & 3 & N/A & 25.83 & 46.31  & 0.50 & 4.29 & 28.05 & 45.59  & 0.48 & 4.33 \\
BMNet+ & 3 & N/A  & 29.47 & 53.15  & 0.51 & 4.72 & 30.74 & 52.00  & 0.47 & 4.68 \\
CounTR  & 3 & N/A  & 21.22 & 40.28  & 0.47 & 3.85 & 21.09 & 40.79  & 0.38 & 3.66 \\
LOCA  & 3 & N/A &  25.70  & 48.64  & 0.45  & 4.30 & 29.93  & 49.89  & 0.48  & 4.62
\\
\midrule
\multicolumn{5}{l}{\textit{Exemplar-free}} \\
CounTR \dag  & 0 & N/A & 23.50 & 45.83 & 0.42 & 4.06 & 23.57 & 42.00 & 0.39 & 3.85 \\
LOCA  & 0 & N/A & 29.37  & 54.01  & 0.51  & 4.84 & 33.96  & 56.66  & 0.53  & 5.17\\
ABC123  & 0 & Max & \textit{19.56} & 46.71 & \textit{0.20} & \textit{3.54} & \textit{22.43} & 47.35 & \textit{0.22} & \textit{3.70}\\
ABC123  & 0 & Sum & \textbf{11.13} & \textbf{34.47} & \textbf{0.12} & \textbf{2.44} & \textbf{11.75} & \textbf{33.41} & \textbf{0.11} & \textbf{2.38}\\
\bottomrule
\end{tabular}

    \label{tab:fsc147_numerical_table}
\end{table}

\subsection{Validating the Number of Predictions}\label{sec:ablation_mhat}

It should be noted that as the matching stage uses the ground-truth density maps, it could be used to significantly benefit a method's quantitative results without improving its deployment capabilities.
Specifically, a method could 
generate a high number of diverse counts and use the matching stage to select the best one. 
We found this to be the case with our method,  see \cref{tab:ablation_mhat} for the complete results.
We believe, however, that this does not align with a more useful method in a deployment situation. This numerical gain derives purely from the matching stage, which is not present during deployment. In fact, during deployment, this would correspond to a much more difficult to interpret output as a user would have to figure out which of the many outputs was most relevant.
We limit ourselves to regressing 5 predictions to minimise this behaviour while still allowing the network to generate valid-but-unknown counts. 
We also found that when high numbers of predictions were generated, fewer than half were used, i.e.\ the outputs of some heads were rarely or never picked. This is likely due to these heads not being matched frequently during training so the loss is rarely propagated back through them. There is also significant redundancy between the heads. The predictions of certain heads over the whole dataset were clearly similar and could be grouped. Of the 39 utilised heads when $\hat{m} = 100$, there were three groups of, respectively, 13, 6, and 4 heads that were very similar, lowering the effective number of utilised heads to 19. 

\begin{table}
    \centering
    \caption{\textbf{Effect of using more prediction heads.} Increasing the number of pre-matching predictions improves the quantitative results of our method. However, as the number of heads increases, the percentage of heads that are frequently ($>0.4\%$) matched decreases.}
    \fontsize{9}{9}\selectfont
    \begin{tabular}{cccccccccccc} 
 \toprule
 & \multicolumn{4}{c}{Val Set} & \multicolumn{4}{c}{Test Set} & \\
 \cmidrule(r){2-5} \cmidrule(r){6-9}

$\hat{m}$ & MAE & RMSE & NAE & SRE & MAE & RMSE & NAE & SRE & Head Utilisation\\ 
\midrule
4 & 9.43 & 17.42 & 0.31 & 2.57 & 10.19 & 19.44 & 0.33 & 2.81 & 100\% \\ 
5 & 8.96 & 15.93 & 0.28 & 2.02 & 9.52 & 17.64 & 0.28 & 2.23 & 100\% \\ 
10 & 8.39 & 14.93 & 0.28 & 1.97 &  9.08 & 16.96 & 0.27 & 2.15 &  100\%\\
20 & 7.78 & 13.75 & 0.26 & 1.82 &  8.29 & 15.53 & 0.24 &  1.95 & 85\%\\ 
50 & 7.26 & 12.80 & 0.25 & 1.69 & 7.99 & 15.14 & 0.24 & 1.81 & 68\%\\ 
100 & 7.11 & 12.81 & 0.23 & 1.59 & 7.43 & 14.48 & 0.21 & 1.72 & 39\%\\ 
 \bottomrule
\end{tabular}
    \label{tab:ablation_mhat}
\end{table}

\section{Conclusion}
\label{conclusion}
In this work, we present ABC123, a multi-class exemplar-free class-agnostic counter, and show that it is superior to prior exemplar-based methods in a multi-class setting.
ABC123 requires no human input at inference-time, works in complex  settings with more than one kind of object present, and outputs easy to understand information in the form of examples of the counted objects.
Due to this, it has potential for deployment in various fields.
We also propose MCAC, a multi-class class-agnostic counting dataset, and use it to train our method as well as to demonstrate that exemplar-based counting methods may not be as robust as previously assumed in multi-class settings. 

% ---- Bibliography ----
%
% BibTeX users should specify bibliography style 'splncs04'.
% References will then be sorted and formatted in the correct style.
%

\clearpage
\clearpage
\clearpage
\clearpage
\bibliographystyle{splncs04}
\bibliography{bib}
\end{document}